\documentclass[10pt,twocolumn,letterpaper]{article}

\usepackage{cvpr}
\usepackage{times}
\usepackage{epsfig}
\usepackage{graphicx}
\usepackage{amsmath}
\usepackage{amssymb}
\usepackage{longtable}
\usepackage{paralist}
\usepackage{multirow,epstopdf,subfigure}
\usepackage{enumitem}

\usepackage[breaklinks=true,bookmarks=false]{hyperref}

\cvprfinalcopy 

\def\cvprPaperID{****} 

\begin{document}

\title{Rain O'er Me: Synthesizing real rain to derain with data distillation}

\author{Huangxing Lin\textsuperscript{1}, \  Yanlong Li\textsuperscript{1}, \  Xinghao Ding\textsuperscript{1*}, \ Weihong Zeng\textsuperscript{1}, \ Yue Huang\textsuperscript{1}, \ John Paisley\textsuperscript{2}\\
\normalsize \textsuperscript{1}School of Information Science and Engineering, Xiamen University, China\\
\normalsize \textsuperscript{2}Department of Electrical Engineering, Columbia University, New York, NY, USA\\
{\small \textsuperscript{*}\tt Corresponding author: dxh@xmu.edu.cn}
}

\maketitle


\begin{abstract}
	We present a supervised technique for learning to remove rain from images without using synthetic rain software. The method is based on a two-stage data distillation approach: 1) A rainy image is first paired with a coarsely derained version using on a simple filtering technique (``rain-to-clean''). 2) Then a clean image is randomly matched with the rainy soft-labeled pair. Through a shared deep neural network, the rain that is removed from the first image is then added to the clean image to generate a second pair (``clean-to-rain''). The neural network simultaneously learns to map both images such that high resolution structure in the clean images can inform the deraining of the rainy images. Demonstrations show that this approach can address those visual characteristics of rain not easily synthesized by software in the usual way.
\end{abstract}

\section{Introduction}

\begin{figure}
	\centering
	\includegraphics[width=0.78in]{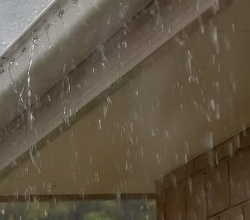}
	\includegraphics[width=0.78in]{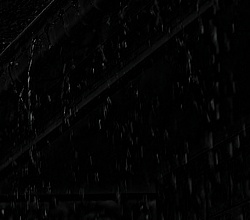}
	\includegraphics[width=0.78in]{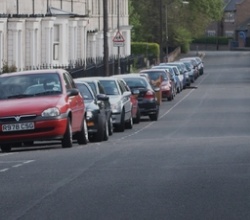}
	\includegraphics[width=0.78in]{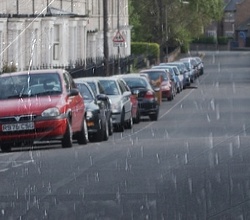}\\
	
	\includegraphics[width=0.78in]{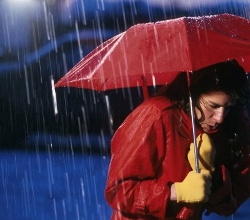}
	\includegraphics[width=0.78in]{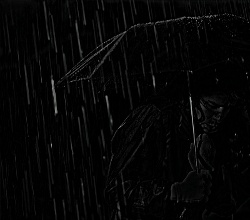}
	\includegraphics[width=0.78in]{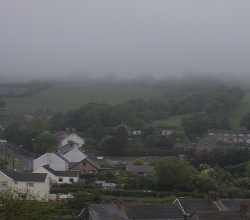}
	\includegraphics[width=0.78in]{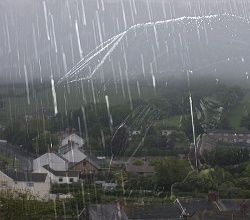}\\
	
	\includegraphics[width=0.78in]{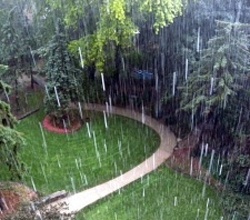}
	\includegraphics[width=0.78in]{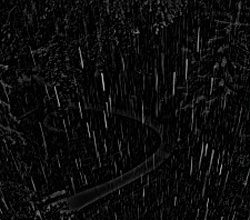}
	\includegraphics[width=0.78in]{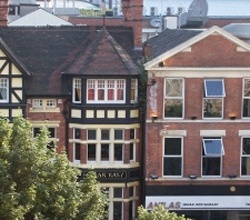}
	\includegraphics[width=0.78in]{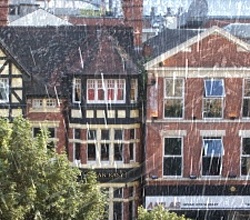}\\
	\vspace{-0.065in}

	\subfigure[]{\includegraphics[width=0.78in]{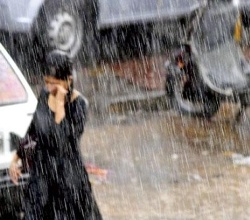}}
	\subfigure[]{\includegraphics[width=0.78in]{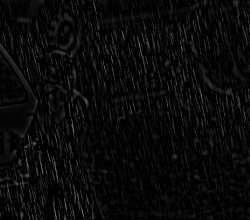}}
	\subfigure[]{\includegraphics[width=0.78in]{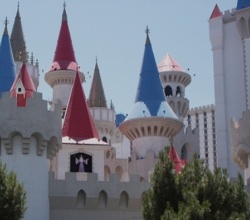}}
	\subfigure[]{\includegraphics[width=0.78in]{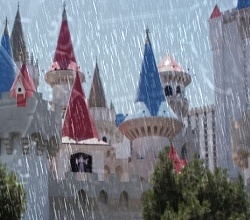}}
	\caption{(a) real rainy image. (b) rain map $\hat{R}_e$ corresponding to (a) generated by our method. (c) real clean image $C$. (d) rainy image pair generated for training, $D=C+\hat{R}_e$.}
	\label{fig1}
\end{figure}

Outdoor vision systems, such as road surveillance, can be negatively impacted by rainy weather conditions.
Many fully-supervised convolution neural networks have been proposed to address this rain removal problem at the single-image level \cite{Yang, fu2017removing, He2018Density, li2018recurrent, Eigen2014Restoring}. These methods use large number of image pairs with and without rain for training, for software is used to synthesize rain in a clean image. While performance is often very good, generalization can be poor when the appearance and style of synthetic rain is different from the real rain. In Figure \ref{fig2} below, we see that synthetic rain tends to be more homogeneous in shape, brightness and direction, while the distribution of real rain is much more irregular. The result is that a model trained with synthetic rain has difficulty in many realistic scenarios. Instead, adding real synthetic rain to a clean image as in Figure \ref{fig1}, a model can more easily learn to recognize and remove the realistic looking object.

At present, the major problem of single image rain removal is improving generalization performance. The ideal solution is to train the networks with only real-world images. Unfortunately, collecting clean/rainy versions of the exact same image is effectively impossible. Another approach is to treat rain removal as an unsupervised learning problem. Some unsupervised methods, such as CycleGAN \cite{zhu2017unpaired} and DualGAN \cite{yi2017dualgan}, have succeeded in cross-domain image translation, but since rain streaks are fairly sparse, these unsupervised methods tend to focus on high energy content in the absence of supervised constraints, thus failing at this problem. (See Figure \ref{fig8}.)

To address this problem, we focus on combining information in real rainy and clean images to mutually aid the deraining process. This is based on a two-stage data distillation method that attempts to perform deraining of both real rainy images and clean images to which the rain extracted from the real images has been added. Like previous knowledge distillation methods \cite{Hinton2015Distilling, zhang2018deep, radosavovic2018data}, our method also creates soft and hard objectives to train this single network. However, our method does not require a strong pre-trained teacher network or a large amount of paired data. 

Instead, we observe that rain is a form of sparse noise that can be suppressed using general image transformation techniques such as scaling and filtering. Therefore, it is easy to remove rain from an image with these basic techniques, but unfortunately much informative image content is removed as well due to over-smoothing, which can have just as negative an impact on any downstream applications. In this paper, we view a derain network as a data distillator, which can distill the noise information (\emph{i.e.} rain streaks) from the input rainy image to help generate new rainy-clean image pairs by adding the removed rain to a clean image. By training a neural network on both the ``soft label'' (rain-to-clean) and ``hard label'' (clean-to-rain) images, we can simultaneously learn to preserve high resolution detail from the latter, while learning to detect and remove realistic looking rain from the former. Our experiments show that our method performs well in nearly any 
real rainy scenario, while the robustness of other state-of-the-art derain methods to non-uniform rain is often more disappointing.

\begin{figure}
	\centering
	\includegraphics[width=0.78in]{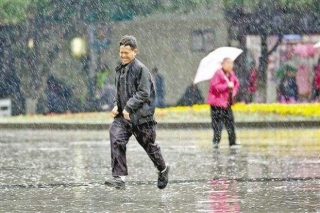}
	\includegraphics[width=0.78in]{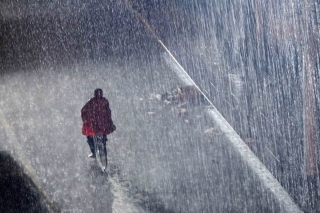}
	\includegraphics[width=0.78in]{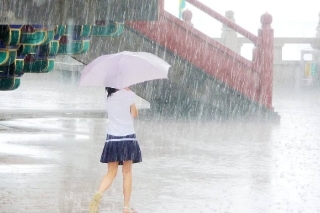}
	\includegraphics[width=0.78in]{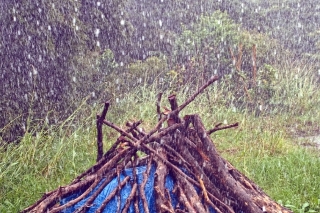}
	
	\subfigure[]{\includegraphics[width=0.78in]{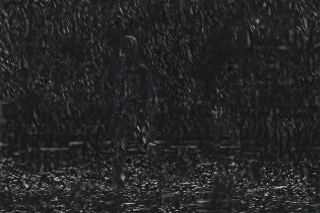}}
	\subfigure[]{\includegraphics[width=0.78in]{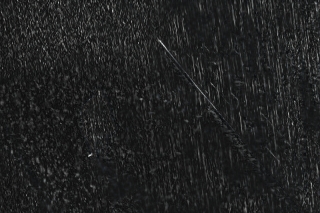}}
	\subfigure[]{\includegraphics[width=0.78in]{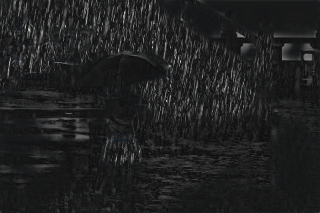}}
	\subfigure[]{\includegraphics[width=0.78in]{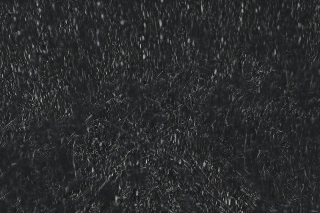}}
	
	\includegraphics[width=0.78in]{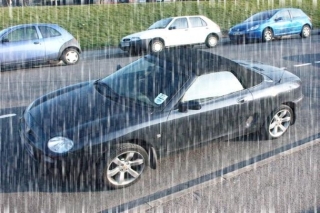}
	\includegraphics[width=0.78in]{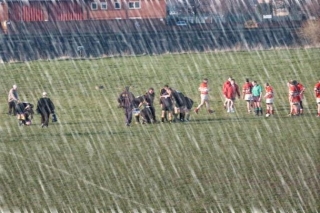}
	\includegraphics[width=0.78in]{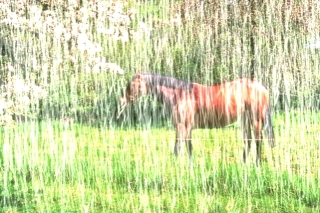}
	\includegraphics[width=0.78in]{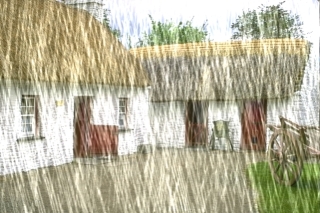}
	
	\subfigure[]{\includegraphics[width=0.78in]{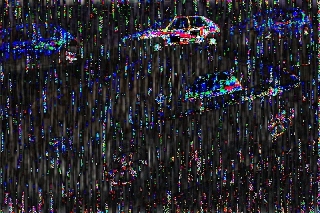}}
	\subfigure[]{\includegraphics[width=0.78in]{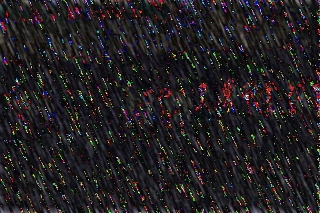}}
	\subfigure[]{\includegraphics[width=0.78in]{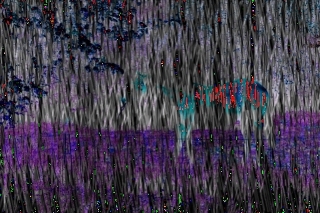}}
	\subfigure[]{\includegraphics[width=0.78in]{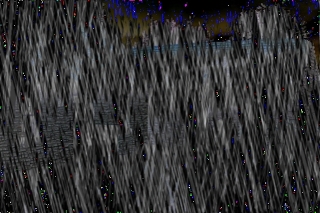}}
	\caption{Visual comparisons of real rainy images and synthesized rainy images. The 1st and 2nd rows: real rainy images from Internet and their corresponding rain maps $\hat{R}$ generated by our derain network. The 3rd and 4th rows: synthesized rainy images and their corresponding rain maps obtained by subtracting ground-truth.}
	\label{fig2}
\end{figure}

\section{Related work}
\paragraph{Single image rain removal.}
Single image de-raining is a challenging and ill-posed task. Traditional methods are designed by using handcrafted image features to describe physical characteristics of rain streaks, or exploring prior knowledge to make the problem easier. Kang \emph{et al.} \cite{Kang2012Automatic} attempt to separate rain streaks from the high frequency layer by sparse coding. In \cite{Chen2013A, Yi2017Transformed}, low-rank assumptions are used to model and separate rain streaks. Kim \emph{et al.} \cite{Kim2014Single} first detect rain streaks and then remove them with the nonlocal mean filter. Luo \emph{et al.} \cite{Yu2015Removing} propose a framework to rain removal based on discriminative sparse coding. Li \emph{et al.} \cite{Li2016Rain} exploit Gaussian mixture models to separate the rain streaks. A limitation of many of these methods is that they tend to over-smooth the resulting image output \cite{Kang2012Automatic, Li2016Rain}.

Recently, deep learning has sped up the progress of single image deraining. In \cite{fu2017removing}, a deep network takes the image detail layer as its input and predicts the negative residual as output. In \cite{Yang}, a recurrent dilated network with multi-task learning is proposed for joint rain streaks detection and removal. In \cite{He2018Density}, Zhang \emph{et al.} propose a density-aware multi-stream densely connected convolutional neural network (DID-MDN) for joint rain density estimation and rain removal. These methods learn a mapping between synthesized rainy images and their corresponding ground truths. A major drawback however, is that this can lead to poor generalization ability to real rainy images that are not easily synthesized for training.

\paragraph{Knowledge distillation.}
Knowledge distillation has been explored to transfer knowledge between varying-capacity networks for supervised modeling \cite{Hinton2015Distilling, Yim2017A, romero2014fitnets,zhang2018deep, ba2014deep}. Hinton \emph{et al.} \cite{Hinton2015Distilling} distilled knowledge from a large pre-trained model to improve a target net, which allowed for additional supervision during training. Radosavovic \emph{et al.} \cite{radosavovic2018data} propose data distillation, which ensembles a model run on different transformations of an unlabeled input image to improve the performance of the target model, but which cannot adapt to unsupervised tasks. Unlike from \cite{radosavovic2018data}, our work focuses on distilling knowledge from the input data to construct extra supervision information in the absence of paired data without using a pre-trained teacher network.


\begin{figure*}
	\centering
	\subfigure[Coarse rain removal to generate the soft label pair.]{\includegraphics[width=7in]{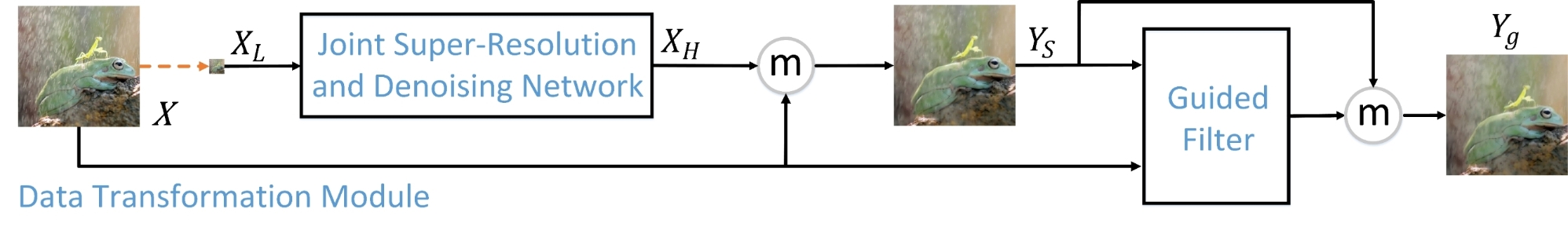}\label{fig.3_a}}
	\subfigure[Two-stage data distillation for deraining. The network uses (a) and a random clean image pair to guide the network.]{\includegraphics[width=7in]{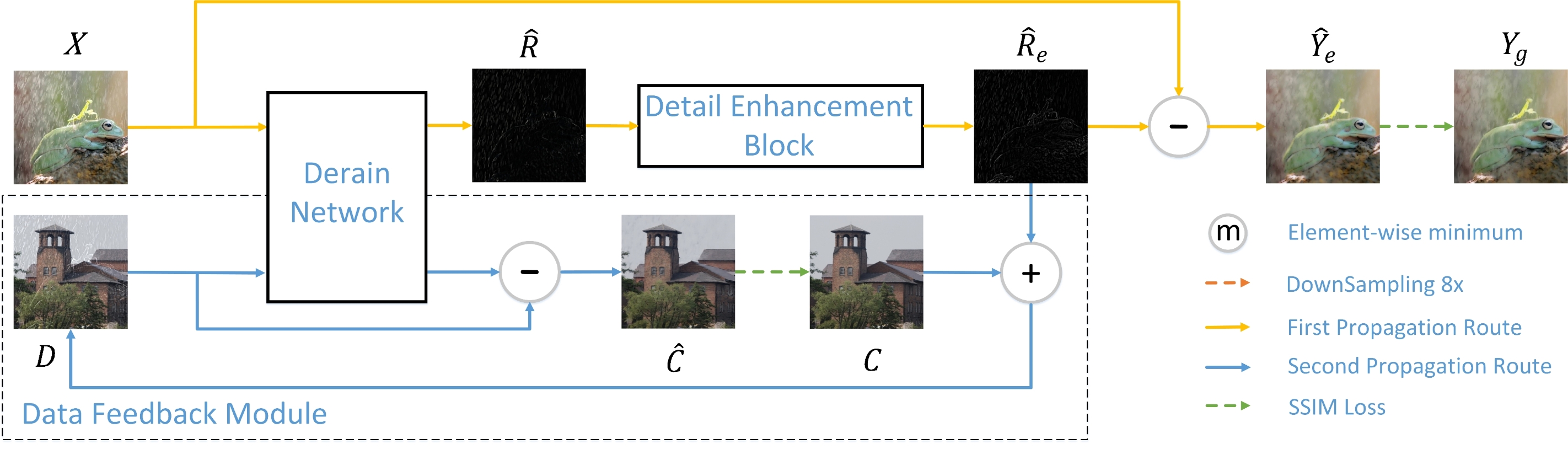}\label{fig.3_b}}
	\caption{The framework of our two-stage data distillation for singal image rain removal. (top) First, a rough deraining algorithm creates a soft label for supervision. The result is an image with much of the rain removed, but significant smoothing. (bottom) We then use a deep network to learn how to remove the rain from the original image while preserving edge details. This is achieved by pairing a rainy image with a different clean image. The derain network then is simultaneously responsible for deraining the true image, and the clean image to which the removed rain has been added. The network thereby learns to remove realistic looking rain, rather than synthetic rain.}
	\label{fig3}
\end{figure*}

\section{Proposed method}
A rainy image $X$ is often considered as a linear combination of a rain-free background $B$ and a rain-streak component (\emph{i.e.} residual map) $R$ \cite{zhu2017joint, liu2018erase},
$X=B+R$. Given $X$ the goal of image deraining is to estimate $B$, which can be equivalently done by estimating the rain residual $R$.

Our goal is to train a derain network using real rainy images without the corresponding clean labels. To generate the necessary supervision information, we propose a two-stage data distillation method as shown in Figure \ref{fig3}. We call this method two-stage because we distill knowledge from the input data twice to form soft and hard supervision. 
\begin{itemize}
	\item In the first stage, the rainy image passes through a predefined data transformation module that easily removes much of the rain streaks, but along with much of the other high resolution as well. This generates a rainless ``soft label'' for the true rainy image to help guide the deraining process. This procedure, while being predefined, is not something that requires learning.
	\item In the second stage, we use a derain network as the data distillator to remove the rain from the true rainy image and add it to a different clean image to generate a new ``hard labeled'' image pair.
\end{itemize}

Under the guidance of soft rain-to-clean objective, the network will learn to remove rain streaks, while the hard clean-to-rain objective will force the network to learn to output images with structural detail. Combining soft and hard tasks in one learning objective, our network learns to output high-quality rain-free images.

\subsection{Data transformation}
Because we don't have real rainy/rain-free image pairs, we cannot use a pre-trained teacher network to generate supervision, as is the case with previous knowledge distillation methods \cite{Hinton2015Distilling, Yim2017A, romero2014fitnets, radosavovic2018data}. Fortunately, we can use some image prior knowledge to generate extra information for learning. For example, we know that rain streaks can be significantly suppressed by some basic image transformation techniques, such as scaling and filtering. We also know that the pixel value of the rain is larger than that of the surrounding pixels.

\begin{figure}
	\centering
	\subfigure[]{\includegraphics[width=1in]{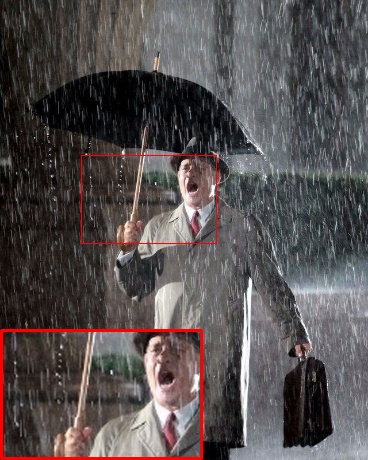}}
	\subfigure[]{\includegraphics[width=1in]{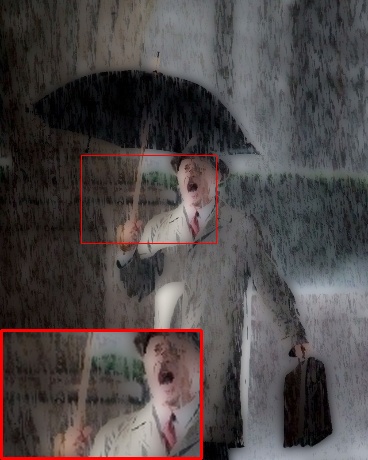}}
	\subfigure[]{\includegraphics[width=1in]{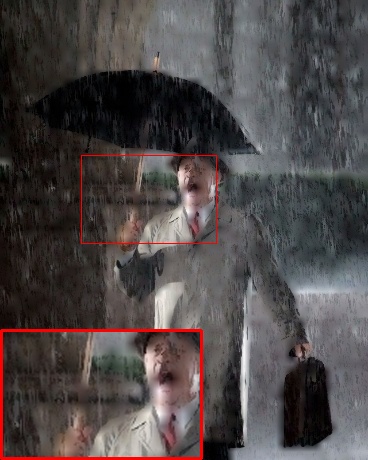}}
	\subfigure[]{\includegraphics[width=1in]{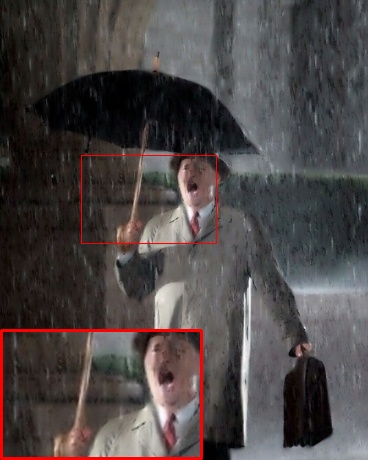}}
	\subfigure[]{\includegraphics[width=1in]{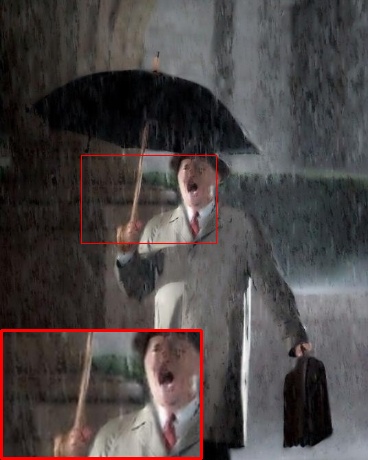}}
	\subfigure[]{\includegraphics[width=1in]{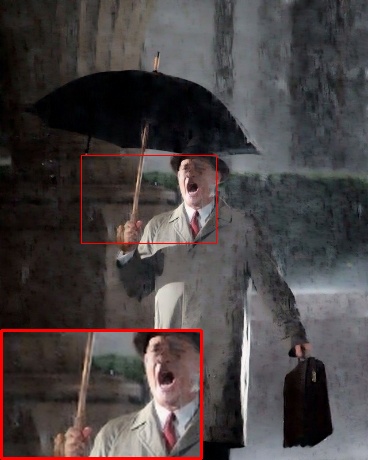}}
	\caption{One example of data transformation using different methods. (a) Rainy image $X$. (b) $Y_S$ generated using low-pass filtering. (c) $Y_S$ generated using Bicubic interpolation. (d) $Y_S$ generated using our scaling operation. (e) $Y_g$, an enhancement of (d) generated using guided filter. (f) Our derained result $\hat{Y}$.}
	\label{fig4}
\end{figure}

Based on these observations, we first build a data transformation module to transform the unlabeled rainy image into a soft label, \emph{i.e.} a rainless but somewhat blurred image. This is shown in Figure \ref{fig.3_a}. The purpose is to use a fast, \textit{unsupervised} algorithm for deraining that can guide the learning. The data transformation module contains two components that are taken as off-the-shelf algorithms: a scaling operation and a filtering operation. We first use Bicubic interpolation to downscale the rainy image $X$ with the scaling factor 8 to generate a low-resolution rainy image $X_L$. Then $X_L$ is returned to its original size $X_H$ via a joint Super-Resolution and Denoising Network (SRDN) based on the deep back-projection network (DBPN) \cite{haris2018deep}. 

Since the rain is sparse, thin and can be viewed as noise, this information will inevitably be lost during scaling the input image along with structural information. To enhance these image details, we calculate the element-wise minimum between $X$ and $X_H$ to produce a relatively clear rainless image $Y_S$, since rain, being on the white end of the spectrum, statistically has larger pixel values. 
We pre-trained the SRDN with unlabeled real rainy images collected from the Internet. The loss function for SRDN can be written as
\begin{equation}
L_{SR}=\frac{1}{M}\sum_{i=1}^{M}\Big(L_{MSE}(X_{H}^i,X^i)+L_{MSE}(X_{H}^i,Y_{S}^i)\Big),
\end{equation}
where $M$ is the number of training data, $L_{MSE}$ denotes MSE loss. $L_{MSE}(X_{H}^i,X^i)$ is a general super resolution loss function. The loss $L_{MSE}(X_{H}^i,Y_{S}^i)$ encourages the SRDN to learn to remove the residual rain in the process of super-resolution. In addition, other related methods, such as low-pass filtering and image interpolation, do not produce better results than our scaling operation. When the scaling operation is replaced by low-pass filtering, the generated images are too blurred, and if we replace SRDN with the Bicubic interpolation, the resulting images will leave obvious artifacts as shown in Figure \ref{fig4}.

When training the derain network discussed later, we fix the parameter weights of SRDN and use it as an image filter. The proposed scaling operation can suppress most of the rain in the inupt rainy image $X$, but it still leaves some rain streaks in the result $Y_S$. To obtain a clearer transformation of $X$, we use a guided filter to further process $X$ and $Y_S$. The guided filter, proposed in \cite{he2010guided} and \cite{he2015fast}, transfers the structures of a guidance image to the filtered output while preserving edges. Since the rainless image $Y_S$ retains most of the structured information of $X$, $Y_S$ can be used as a guided image and convolved with the input rainy image $X$ via the guided filter to obtain the final rain-free image $Y_g$, as shown in Figure \ref{fig4}. $Y_g$ can be viewed as a rain-free soft label for the rainy image $X$. We call it a ``soft'' label, because some details of the background are also lost during this data transformation.

\begin{figure}
	\centering
	\includegraphics[width=3.5in]{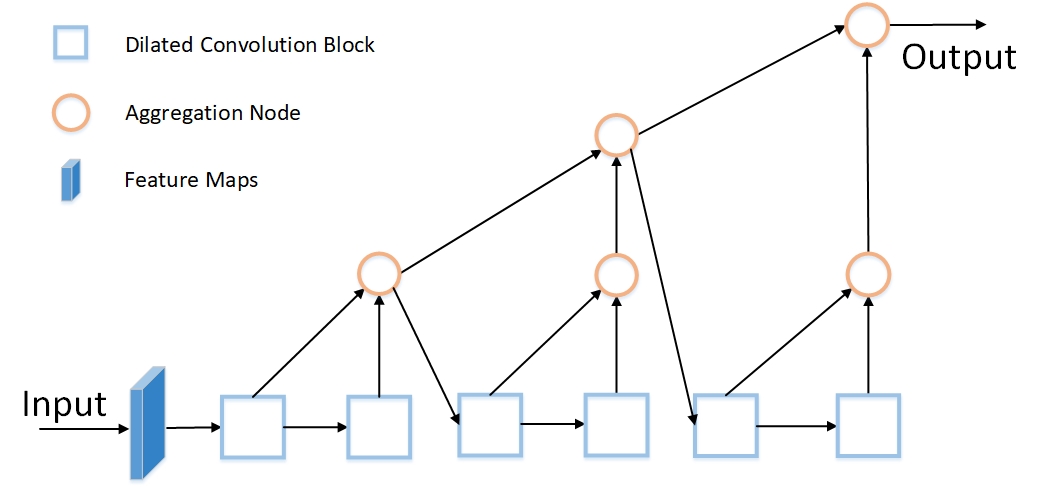}
	\caption{The hierarchical aggregation architecture of our derain network. Hierarchical aggregation learns to extract the full spectrum of semantic and spatial information from the network. The derain network contains six dialted blocks and the number of feature maps is 16 for all convolution layers.}
	\label{fig5}
\end{figure}
\begin{figure}
	\centering
	\includegraphics[width=3in]{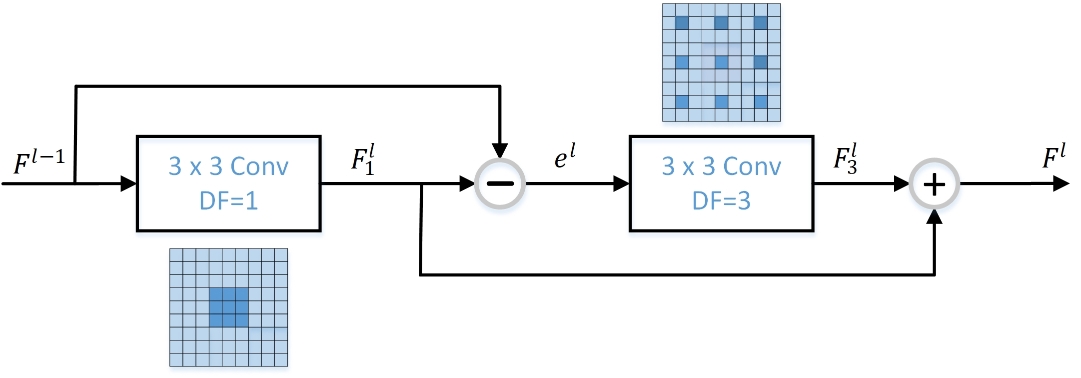}
	\caption{The structure of the dilated convolution block. DF indicates the dilated factor.}
	\label{fig6}
\end{figure}

\subsection{Derain network}
Many deep convolutional networks have been designed to handle the single image rain removal \cite{Yang, fu2017removing, He2018Density, li2018recurrent}. Most existing deep methods design a very complex network in order to obtain higher numerical performance on synthetic datasets, but at the cost of some poor generalization, scalability and practicality in real-world image applications.


Instead of directly cascading convolutional layers, we design a hierarchical aggregation architecture, as shown in Figure \ref{fig5}, to better fuse spatial and semantic information across layers \cite{yu2018deep}, which can lead to high quality images reconstruction. We argue that effectively aggregating features can lead to better rain removal as well as better parameter and memory efficiency.
On the other hand, unlike the usual noise, the appearance of the rain streak is irregular, as shown in Figure \ref{fig2}. To better capture the characteristics of rain, we design the multi-scale dilated convolution block shown in Figure \ref{fig6} as the backbone of the derain network. The block is defined as follows,
\begin{equation}
\begin{aligned}
&F_{1}^l=W_{1}^l*F^{l-1},\\
&e^l=F^{l-1}-F_{1}^l,\\
&F_{3}^l=W_{3}^l*e^l,\\
&F^l=F_{1}^l+F_{3}^l,
\end{aligned}
\end{equation}
where $F$ and $W$ are feature maps and $3\times3$ convolution kernels, respectively. The subscript number are dilation factors, $*$ indicates the convolution operation, $l$ indexes block number.
The multi-scale dilated convolution block can also be viewed as a self-correcting procedure that feeds a mapping error to the sampling layer at another scale and iteratively corrects the solution by feeding back the mapping error. 

Moreover, the features between blocks are fused by the aggregation nodes. The nodes learn to select and project important information to maintain the same dimension at their output as a single input. In the derain network, all the nodes use $3\times3$ convolutions. The activation function for all convolutional layers is $ReLU$ \cite{krizhevsky2012imagenet}, while the activation function of the last layer is $Tanh$. To ease learning, the direct output of the derain network is the residual map $\hat{R}$,
\begin{equation}
\hat{R}=f(X,\theta),
\end{equation}
where $f$ denotes the mapping function of the derain network, $\theta$ represents parameters of the network. The corresponding output $\hat{Y}$ is obtained by 
\begin{equation}
\label{eq.4}
\hat{Y}=X-\hat{R}.
\end{equation}

Then the soft objective for the derain network can be represented as the following loss function
\begin{equation}
\label{eq.7}
L_s=\frac{1}{M}\sum_{i=1}^{M}L_{SSIM}(\hat{Y}^i,Y_g^i),
\end{equation}
where $L_{SSIM}$ is the SSIM loss \cite{wang2004image}.


\subsection{Data feedback}
In general, the derain network can learn to remove rain under the supervision of the real rainy images and their corresponding rain-free soft labels. However, the output of the network is somewhat blurred. This is because when the data transformation module distills the soft label from the input image, some details of the image background are also lost. Guided by only the soft objective, the network can learn to remove rain, but can not learn to retain the details of the background. Therefore, additional constraints should be introduced so that the network can learn to preserve the details of the background when removing rain from images.

We find that the derain network can be regarded as a data distillator. The process of the derain network taking in the input image and generating the corresponding residual map $\hat{R}$ can be seen as a distillation of the input image $X$. The residual map (mainly rain streaks) is the noise information of the input rainy image, and we can use this to generate extra knowledge. Specifically, if we add $\hat{R}$ to another clean image $C$ to generate a new rainy image $D$, then we have a new rainy/clean pair where the rainy image has sharp details in both the input and output. The new rainy image $D$ and input image $X$ have the same rain streaks, but their backgrounds are different; both are sharp, but the soft label for $X$ is more blurry while the rainy $D$ is still high resolution. We refer to the new image pair $C$ and $D$ as a hard labeled image pair because the rainy image $D$ has a corresponding clean and sharp label $C$. We can then use the hard labeled image pair while training derain network as well. 


To guide the network to preserve the details of the output images, the corresponding hard objective can represented by the following loss function
\begin{equation}
\label{eq.9}
L_H=\frac{1}{M}\sum_{i=1}^{M}L_{SSIM}(\hat{C}^i,C^i).
\end{equation}

\begin{figure*}
	\centering
	\includegraphics[width=1.1in]{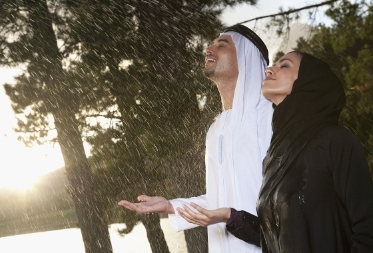}
	\includegraphics[width=1.1in]{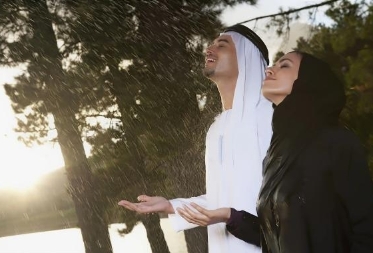}
	\includegraphics[width=1.1in]{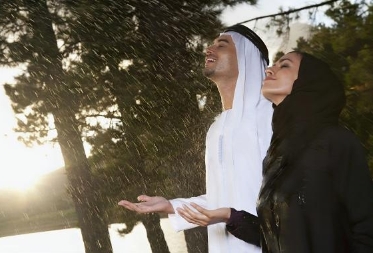}
	\includegraphics[width=1.1in]{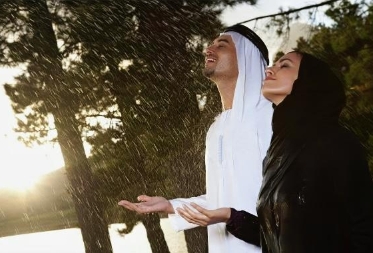}
	\includegraphics[width=1.1in]{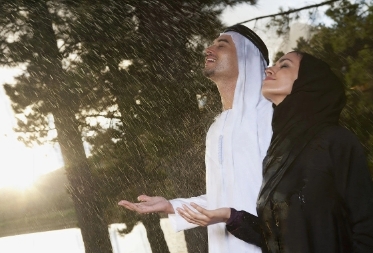}
	\includegraphics[width=1.1in]{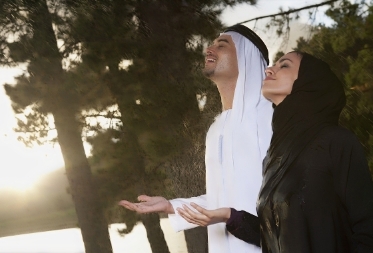}\\
	\includegraphics[width=1.1in]{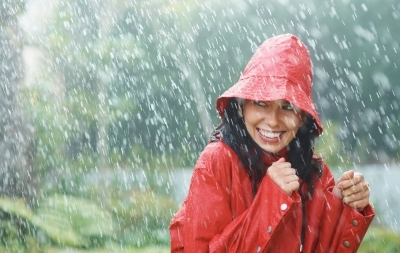}
	\includegraphics[width=1.1in]{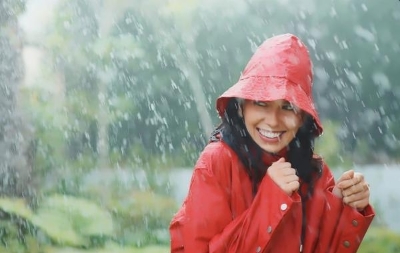}
	\includegraphics[width=1.1in]{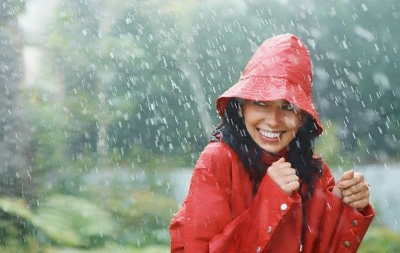}
	\includegraphics[width=1.1in]{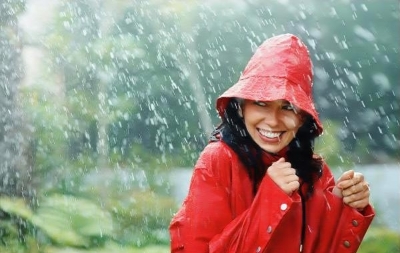}
	\includegraphics[width=1.1in]{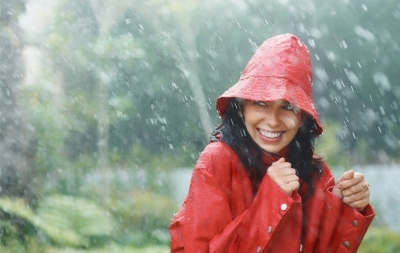}
	\includegraphics[width=1.1in]{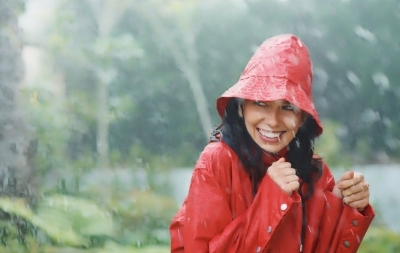}\\
	\includegraphics[width=1.1in]{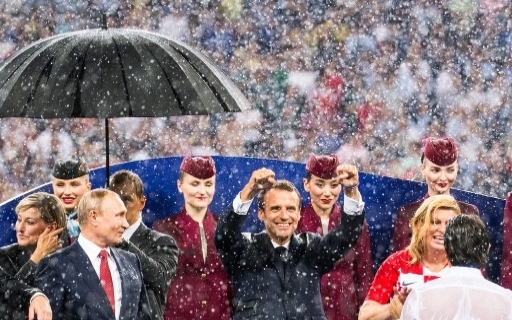}
	\includegraphics[width=1.1in]{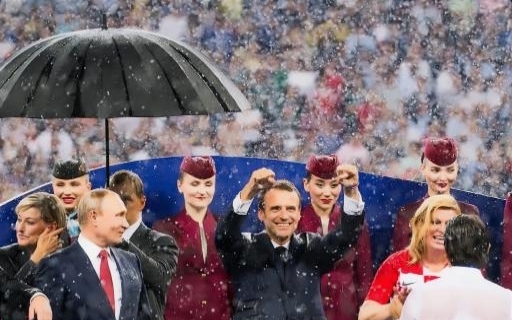}
	\includegraphics[width=1.1in]{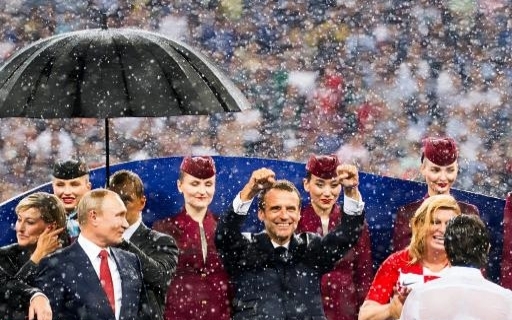}
	\includegraphics[width=1.1in]{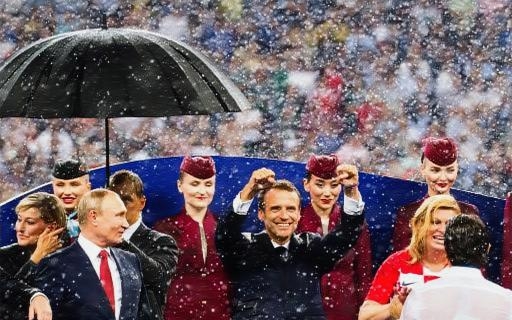}
	\includegraphics[width=1.1in]{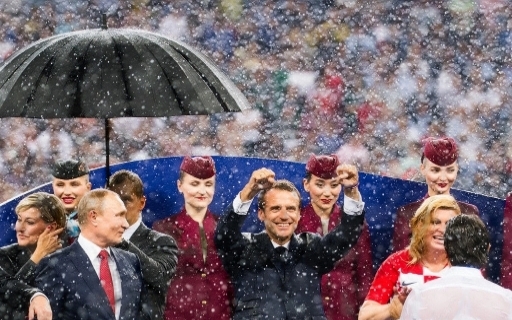}
	\includegraphics[width=1.1in]{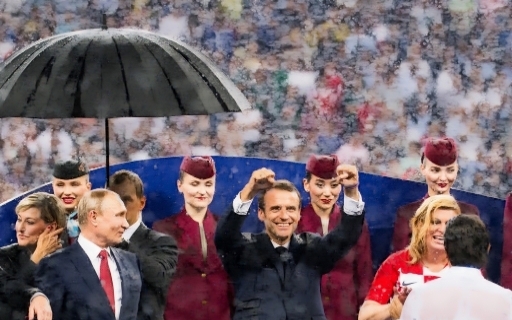}\\
	
	\includegraphics[width=1.1in]{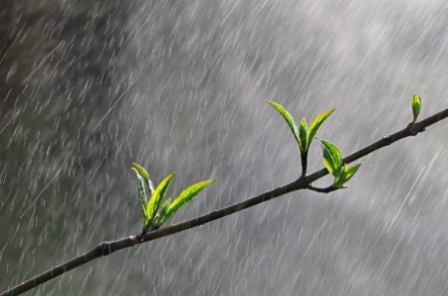}
	\includegraphics[width=1.1in]{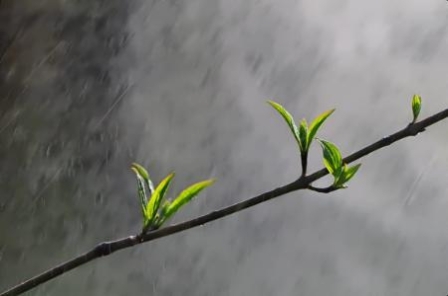}
	\includegraphics[width=1.1in]{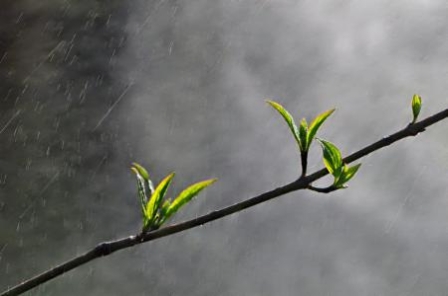}
	\includegraphics[width=1.1in]{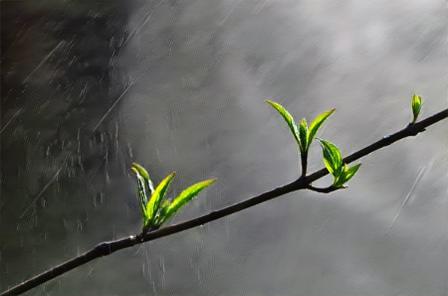}
	\includegraphics[width=1.1in]{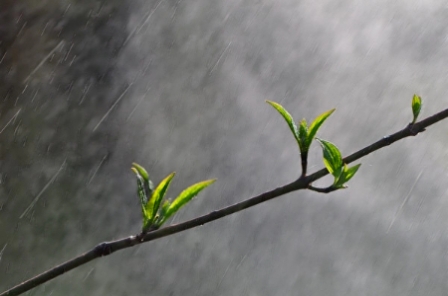}
	\includegraphics[width=1.1in]{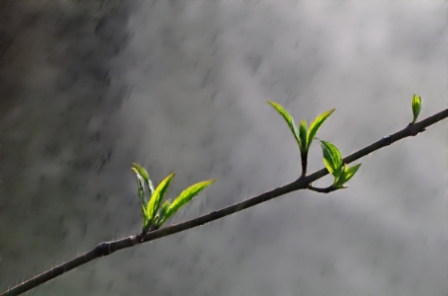}\\
	\includegraphics[width=1.1in]{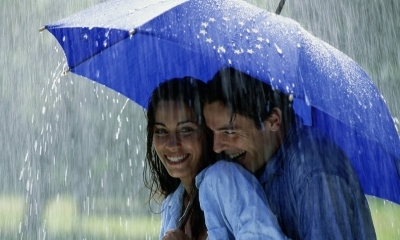}
	\includegraphics[width=1.1in]{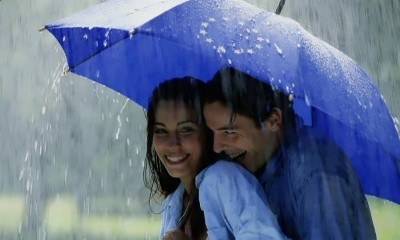}
	\includegraphics[width=1.1in]{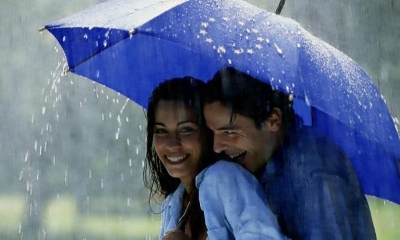}
	\includegraphics[width=1.1in]{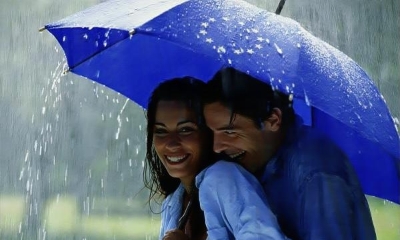}
	\includegraphics[width=1.1in]{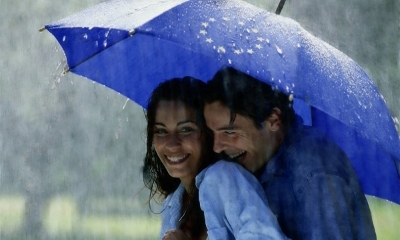}
	\includegraphics[width=1.1in]{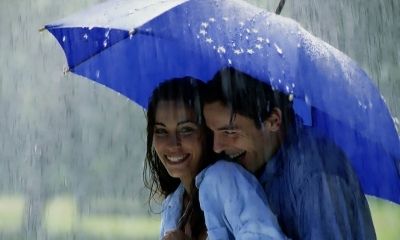}\\
	\includegraphics[width=1.1in]{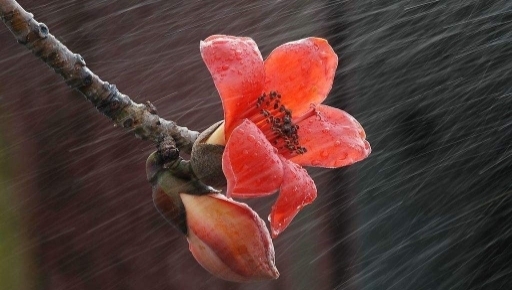}
	\includegraphics[width=1.1in]{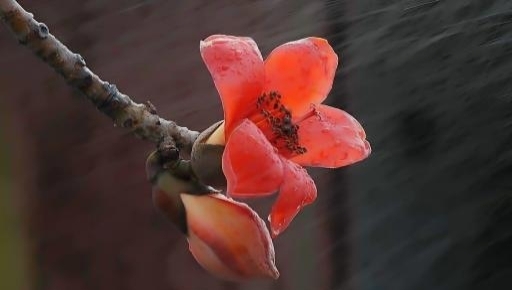}
	\includegraphics[width=1.1in]{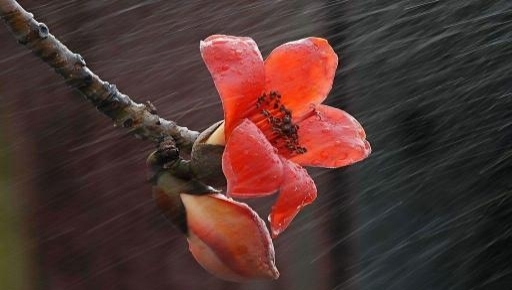}
	\includegraphics[width=1.1in]{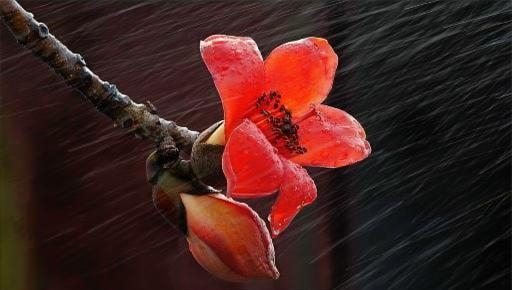}
	\includegraphics[width=1.1in]{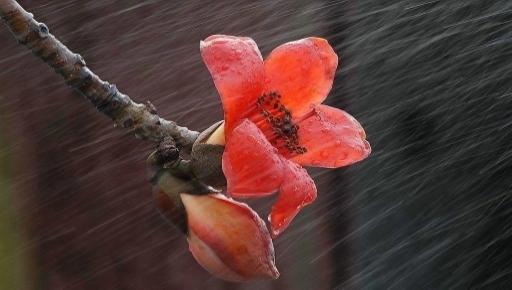}
	\includegraphics[width=1.1in]{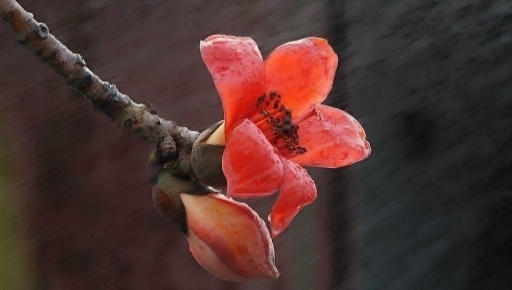}\\
	\vspace{-0.065in}
	\subfigure[Rainy images]{\includegraphics[width=1.1in]{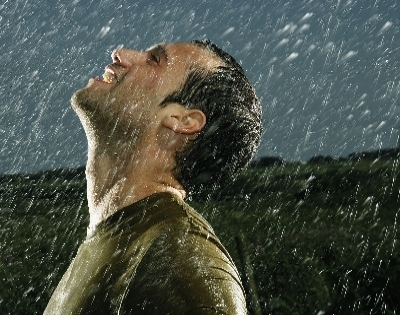}}
	\subfigure[GMM]{\includegraphics[width=1.1in]{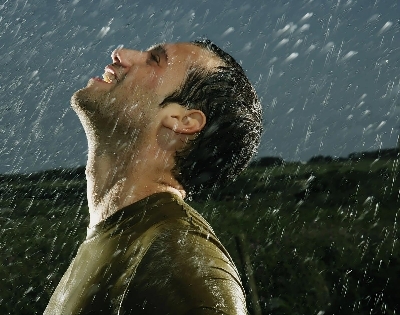}}
	\subfigure[DDN]{\includegraphics[width=1.1in]{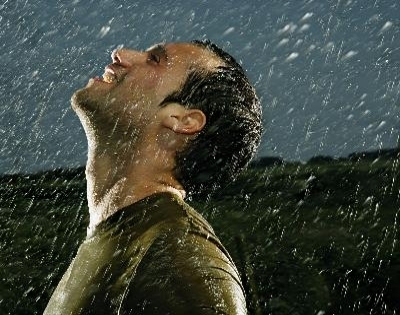}}
	\subfigure[DID-MDN]{\includegraphics[width=1.1in]{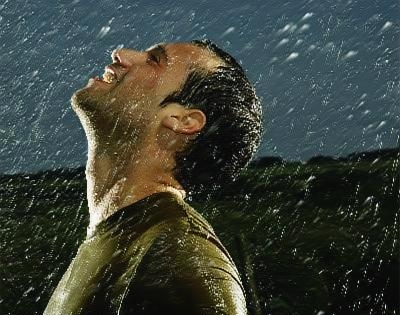}}
	\subfigure[JORDER]{\includegraphics[width=1.1in]{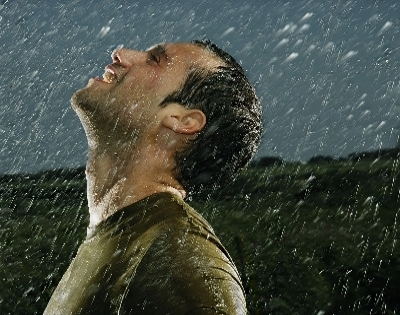}}
	\subfigure[Ours]{\includegraphics[width=1.1in]{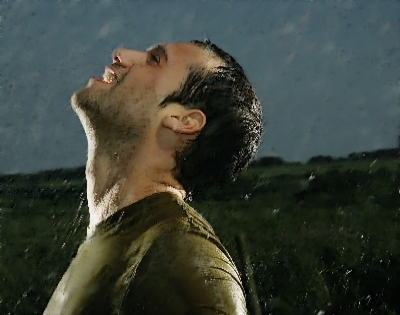}}
	\caption{Visual comparisons on real rainy images. We have chosen images that do not have typical look of synthetic rain and which would be difficult to generate using current software. In many cases, the existing algorithms can remove rain that is homogenous and synthetic-like. However, when this is not the case our algorithm still removes the rain while others fail. This is because data distillation allows our network to be trained on rain with this appearance, while the other algorithms are not.}
	\label{fig7}
\end{figure*} 

\begin{figure}
	\centering
	\subfigure[Rainy image]{\includegraphics[width=1.6in]{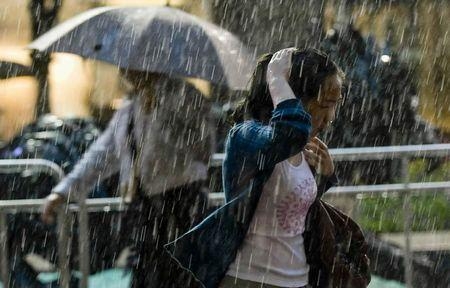}}
	\subfigure[CycleGAN]{\includegraphics[width=1.6in]{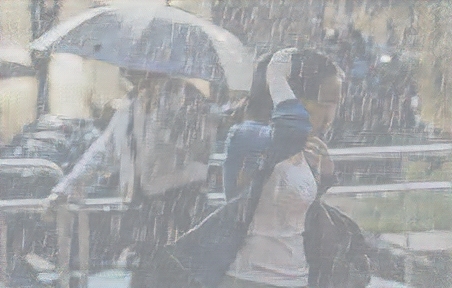}}
	\subfigure[DualGAN]{\includegraphics[width=1.6in]{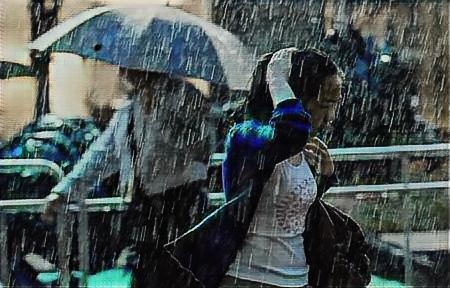}}
	\subfigure[Our result]{\includegraphics[width=1.6in]{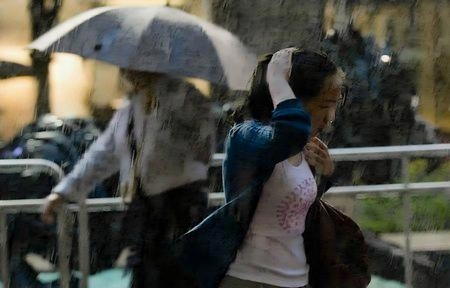}}
	\caption{Visual comparisons with unsupervised methods.}
	\label{fig8}
\end{figure}

\begin{figure*}[th!]
	\centering
	\includegraphics[width=1.1in]{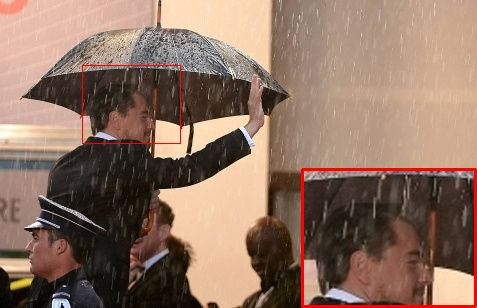}
	\includegraphics[width=1.1in]{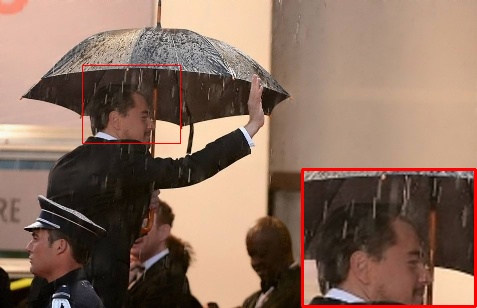}
	\includegraphics[width=1.1in]{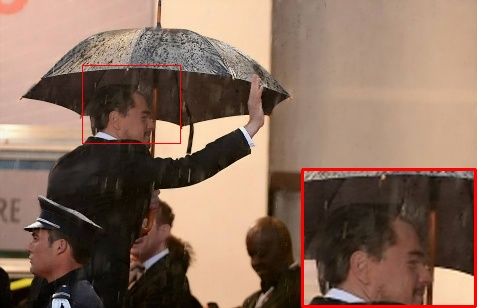}
	\includegraphics[width=1.1in]{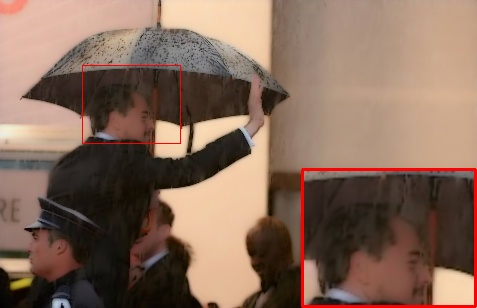}
	\includegraphics[width=1.1in]{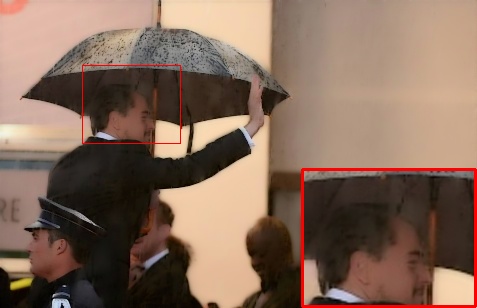}
	\includegraphics[width=1.1in]{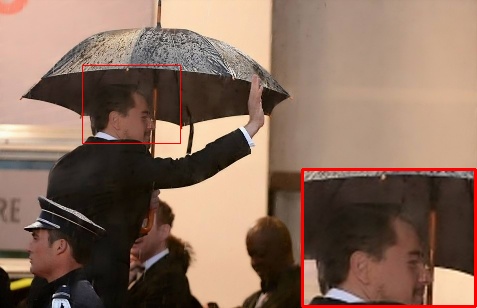}\\
	
	\subfigure[Rainy images]{\includegraphics[width=1.1in]{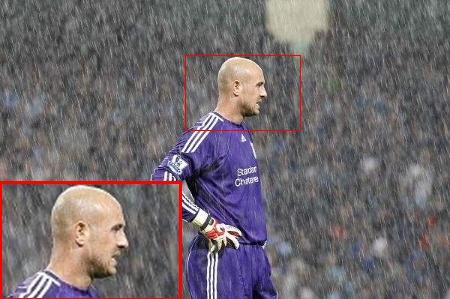}}
	\subfigure[No-scaling]{\includegraphics[width=1.1in]{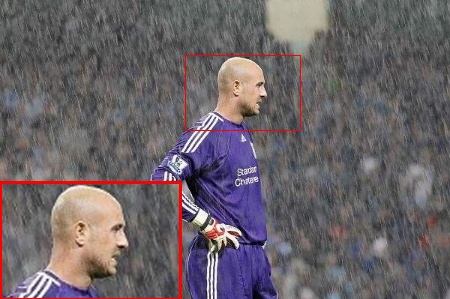}}
	\subfigure[No-filter]{\includegraphics[width=1.1in]{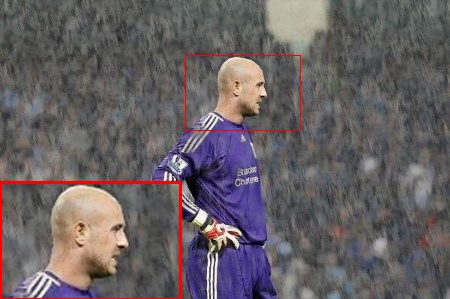}}
	\subfigure[No-feedback]{\includegraphics[width=1.1in]{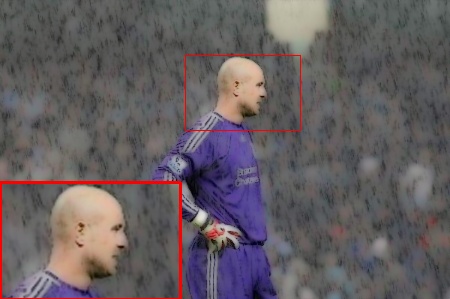}}
	\subfigure[No-detail]{\includegraphics[width=1.1in]{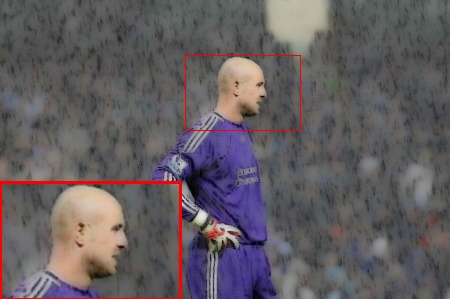}}
	\subfigure[Ours]{\includegraphics[width=1.1in]{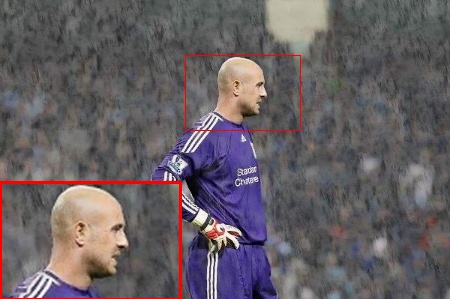}}
	\caption{Visual comparisons on ablation study.}
	\label{ablation}
\end{figure*} 

\begin{figure}
	\centering
	\includegraphics[width=1.6in]{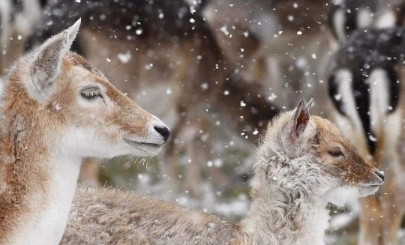}
	\includegraphics[width=1.6in]{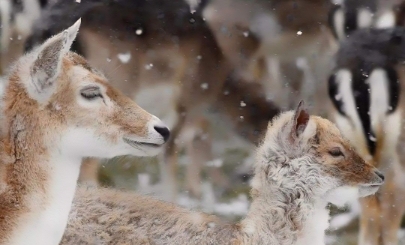}
	\subfigure[Real snow images]{\includegraphics[width=1.6in]{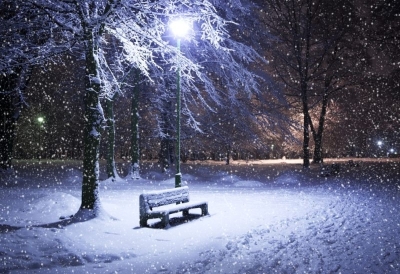}}
	\subfigure[Our results]{\includegraphics[width=1.6in]{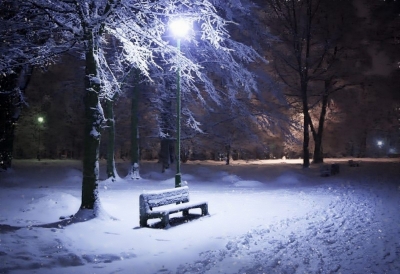}}
	\caption{Extension on single image snow removal.}
	\label{desnow}
\end{figure}

\begin{figure}
	\centering
	\includegraphics[width=0.78in]{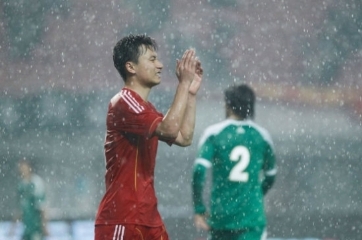}
	\includegraphics[width=0.78in]{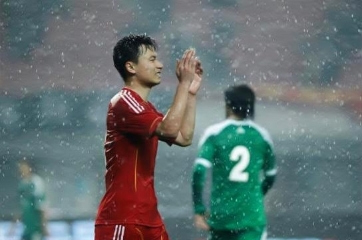}
	\includegraphics[width=0.78in]{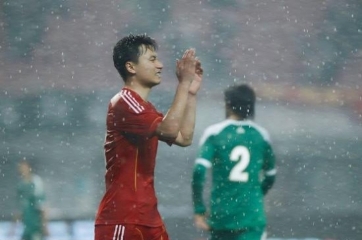}
	\includegraphics[width=0.78in]{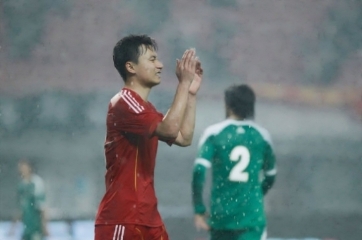}\\
	
	\includegraphics[width=0.78in]{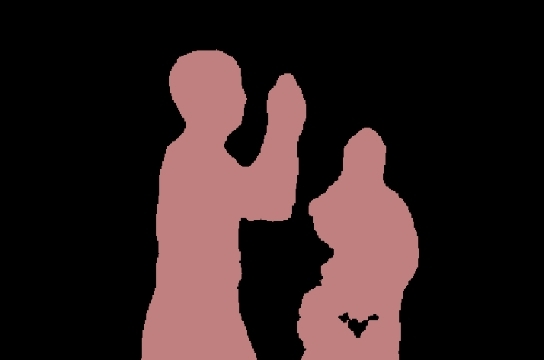}
	\includegraphics[width=0.78in]{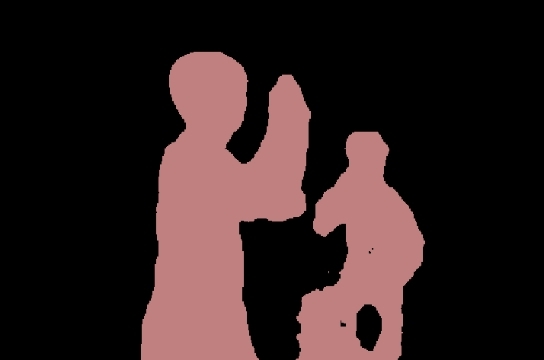}
	\includegraphics[width=0.78in]{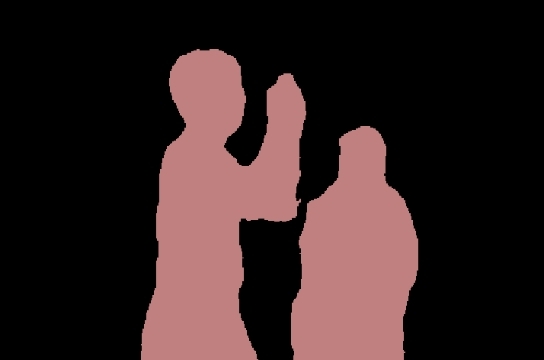}
	\includegraphics[width=0.78in]{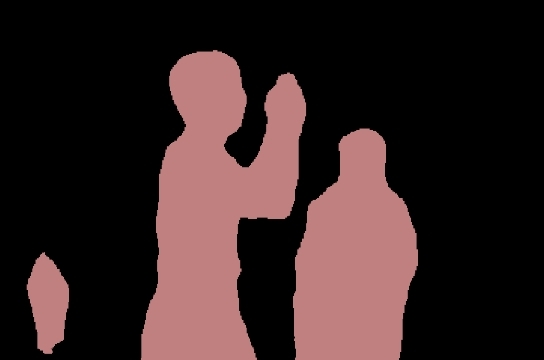}\\
	
	\includegraphics[width=0.78in]{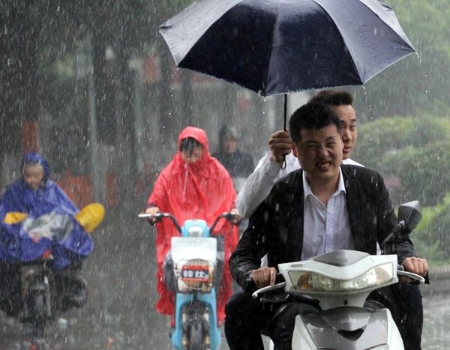}
	\includegraphics[width=0.78in]{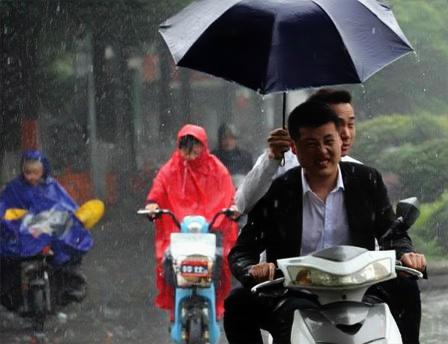}
	\includegraphics[width=0.78in]{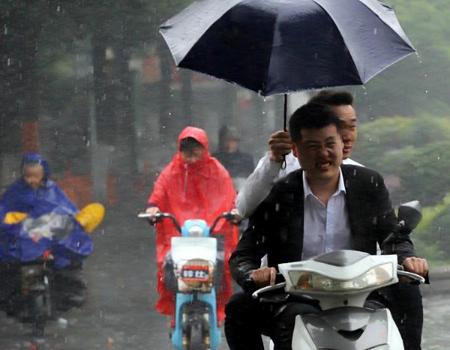}
	\includegraphics[width=0.78in]{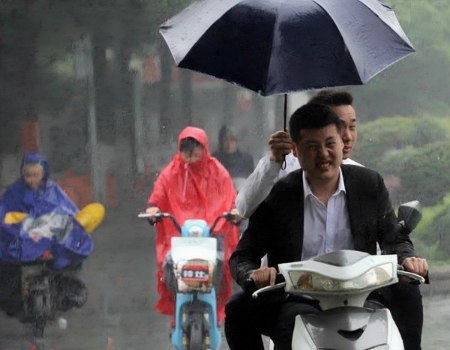}\\
	\vspace{-0.065in}
	
	\subfigure[Rainy images]{\includegraphics[width=0.78in]{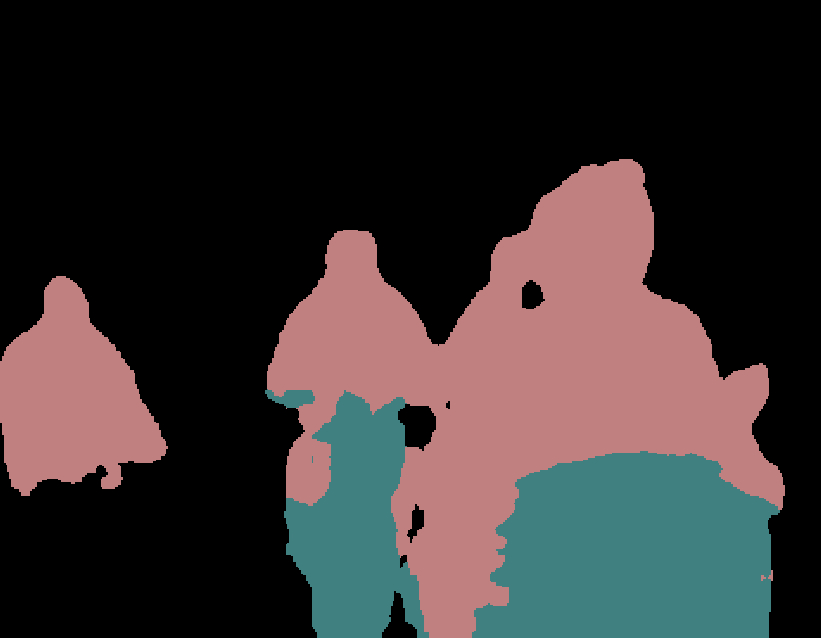}}
	\subfigure[DID-MDN]{\includegraphics[width=0.78in]{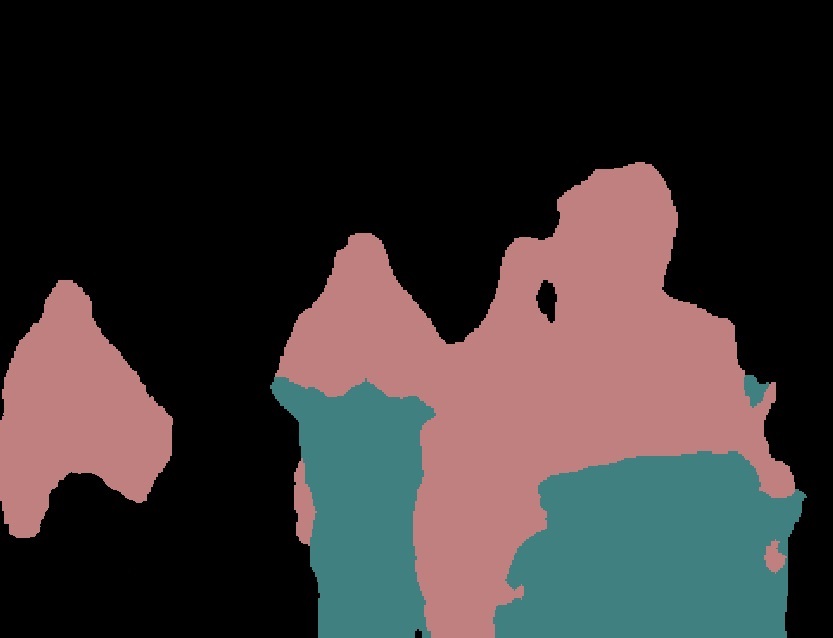}}
	\subfigure[DDN]{\includegraphics[width=0.78in]{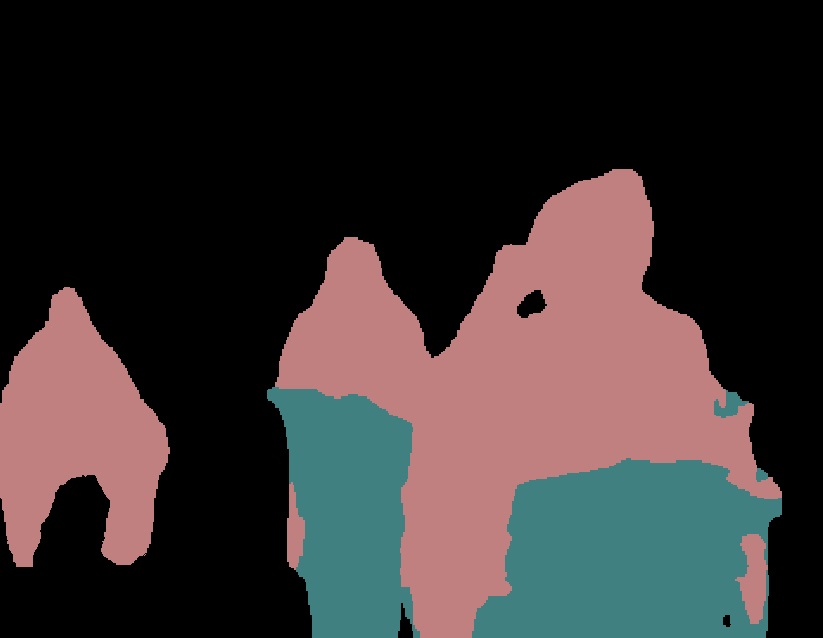}}
	\subfigure[Ours]{\includegraphics[width=0.78in]{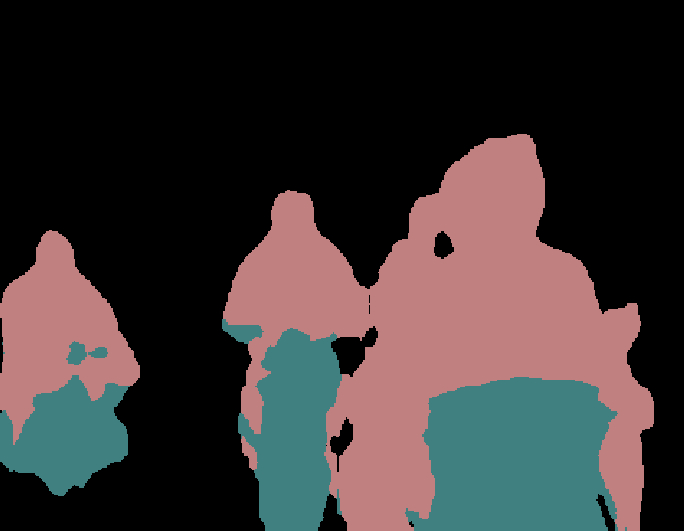}}
	\caption{The 1st and 3rd rows: real rainy images and rain removal results generated by DID-MDN, DDN and our derain network. The 2nd row: the corresponding segmentation maps of the 1st row generated using DeepLabv3+ \cite{chen2018encoder}. The 4th row: the corresponding segmentation maps of the 3rd row.}
	\label{segmentation}
\end{figure}

\subsection{Detail enhancement}

A drawback of objective (\ref{eq.7}) is that the derain network learns to output the rain-free \textit{blurred} soft-label image, while objective (\ref{eq.9}) encourages the network to maintain the details of the clean image. These two objectives are competing. If (\ref{eq.7}) and (\ref{eq.9}) are used directly to train the network, the network produces less than satisfactory results. To address this issue, we introduce an additional detail enhancement block. This block is composed of two $3\times3$ convolution layers. We input the residual map $\hat{R}$ to the detail enhancement block and generate an enhanced residual map $\hat{R}_e$ with richer details. 


Because $\hat{R}_e$ contains more details, the output $\hat{Y}_e = X-\hat{R}_e$ is a rain-free \text{and} blurred image, which is more ideal given the blurriness of the soft-label. The objective now is redefined as
\begin{equation}
\label{eq.12}
L_S=\frac{1}{M}\sum_{i=1}^{M}L_{SSIM}(\hat{Y}_e^i,Y_g^i).
\end{equation}

Because of this detail enhancement block, the derain network only needs to focus on preserving the details of the output and removing the rain, rather than blurring the output to match the soft-label.

We now no longer add $\hat{R}$, but add $\hat{R}_e$ to $C$ to generate the new rainy image $D$. The reason is that we can only ensure that there are rain streaks in $\hat{R}_e$ under the constraint of (\ref{eq.12}), but it is impossible to predict whether there are rain streaks in the feature map $\hat{R}$ of the middle layer. Only there are rain streaks in $D$, can the derain network learn to output rain-free and clean images.

Combining the hard objective (\ref{eq.9}) with the new soft objective (\ref{eq.12}), the complete objective function $L$ is
\begin{equation}
\label{eq.13}
L=L_S+\alpha \cdot L_H,
\end{equation}
where $\alpha$ is the parameter to balance the two losses. We set $\alpha=4$ based on cross validation.

We note that we only use the soft objective function (\ref{eq.12}) to train the network in the first 100 epoch. This means that in the first 100 epochs, the data feedback module does not work. In the next 9900 epochs we use the complete objective function (\ref{eq.13}) to train the network. The reason for this is that in the early stages of training, the output of the network contains too much noise. When the derain network learns to remove rain to some extent,  the data feedback module will work.

Note that when testing, we remove the detail enhancement block and use the derain network directly to get the final output (\ref{eq.4}).


\begin{table*}
	\caption{Comparison of parameters and running time (seconds). The size of the testing images: $512\times512$.}
	\centering
	\begin{tabular}{|c|c|c|c|c|c|c|c|c|c|c|c|}
		\hline
		\multicolumn{2}{|c|}{} & \multicolumn{2}{|c|}{GMM \cite{Li2016Rain}} & \multicolumn{2}{|c|}{DDN \cite{fu2017removing}} & \multicolumn{2}{|c|}{DID-MDN \cite{He2018Density}}& \multicolumn{2}{|c|}{JORDER \cite{Yang}} &    \multicolumn{2}{|c|}{Ours}\\
		\hline
		\multicolumn{2}{|c|}{} &CPU & GPU&CPU & GPU&\ \ CPU \ \ & GPU&CPU & GPU&CPU & GPU \\
		\hline
		\multicolumn{2}{|c|}{Running time} &$1.85\times10^{3}$ & -& $1.56$ & $0.15$ &$5.73$ & $0.16$&$2.77\times10^{2}$ & $0.19$&$1.45$ & $0.15$ \\
		\hline
		\multicolumn{2}{|c|}{Parameters $\#$} &\multicolumn{2}{|c|}{-} & \multicolumn{2}{|c|}{$58,175$}&\multicolumn{2}{|c|}{$135,800$} &\multicolumn{2}{|c|}{$369,792$}  &\multicolumn{2}{|c|}{$51,552$} \\
		\hline
		
	\end{tabular}
	\label{table1}
\end{table*}

\subsection{Training details and parameter settings}
It is hard to obtain rainy/clean image pairs from real-world data, but it is relatively easy to collect a large number of real rainy images. We collect 1600 real rainy images from the Internet, which are diverse in background and rain.\footnote {We will release our code and data.} We divide these images into a training set and a testing set in a radio of 7:1. We first train the SRDN with the training set according to parameter settings described in \cite{haris2018deep}. When training the derain network, we fixed the parameters of SRDN and used it as a component of the data transformation module. We use Pytorch and Adam \cite{kingma2014adam} with a mini-batch size of 8 to train our derain network. We randomly select 256$\times$256 image patchs from training set as inputs. We set the learning rate as $2\times10^{-4}$ for the first 5000 epochs and linearly decay the rate to 0 for the next 5000 epochs. All experiments are performed on a server with Inter Core i7-8700K CPU and NVIDIA GeForce GTX 1080 Ti.

\section{Experiments}
We compare our method with several state-of-the-art supervised derain methods, as well as unsupervised methods. Unlike previous methods, which only pursue higher numerical metrics on \textit{synthetic} data, our desire is devoted to a better qualitative generalization to real-world scenarios.

\subsection{Comparison with derain methods}
We compare our method with the following derain methods in the same test environment: Gaussian Mixture Model (GMM) \cite{Li2016Rain}, DDN \cite{fu2017removing}, DID-MDN \cite{He2018Density} and JORDER \cite{Yang}. Since no ground truth exists, we only show the qualitative results on real-world rainy images in Figure \ref{fig7}. As can be seen, the model-based method GMM fails to remove rain due to the modeling limitations. Among the other deep methods, JORDER has better performance. But overall, the effects of these three fully-supervised methods are disappointing since they cannot remove the real rain that they haven't seen in the synthesized data. On the other hand, our method can remove many types of rain, from small raindrops to long rain streaks, and reconstruct an image that still preserves details. We argue that, compared with other methods this approach is more robust to realistic data.

The specific derain network we use can also process new images very efficiently. Table \ref{table1} shows the average running time of 100 test images, all the test are conducted with a $512\times512$ rainy image as input. The GMM is a non-deep method that is run on CPU according to the provided code, while other deep methods are tested on both CPU and GPU. Compared with other methods, our network has a relatively fast speed on both CPU and GPU. As a pre-processing for other high-level vision tasks, the rain streaks removal process should be simpler and faster. Our derain network is a relatively shallow network that requires fewer calculations, so it is more practical, for example, on mobile devices. It's improved performance is a function of improved training data obtained through our data distillation approach.

\subsection{Comparison with unsupervised methods}
We also compare our method with two unsupervised methods: CycleGAN \cite{zhu2017unpaired} and DualGAN \cite{yi2017dualgan}. Both GAN-based models are good at domain transfer, but they fail to transfer the rainy image into a rain-free image as shown in Figure \ref{fig8}. The main reason is that rain is a sparse and low-energy component of the image, and these unsupervised methods can only capture salient characteristics of the image in the absence of supervised constraints. Therefore, they tend to ignore the difficult rain removal problem but focus on the easy style transfer problem. In contrast, our method still works well for rain removal in the absence of ground truth training data pairs, again because we use data distillation to create ``synthetic'' data only using content extracted from real images.


\subsection{Ablation study}
%
To validate the necessity of scaling operation (\emph{i.e.} downscale and super-resolution), guided filter, data feedback module and detail enhancement block, we design four variants of the two-stage data distillation method. One is called No-scaling, which means we remove the scaling operation. The second is called No-filter, which removes the guided filter. The third, called No-feedback, removes the data feedback module. The last one is call No-detail, which removes the detail enhancement block. We again abbreviate the complete two-stage data distillation method as “Ours”.

Subjective comparisons are presented in Figure \ref{ablation}. As can be seen, without scaling, many rain streaks remain in the output image. Without data feedback module, the network generates over-smoothed image. Without the detail enhancement block, the network can also remove rain, but the results are somewhat blurred. The guided filter is helpful to the rain removal performance of the network as well.

\subsection{Extensions}
Interestingly, we find that our derain network can be applied directly to image snow removal without having to be retrained with another dataset, as shown in Figure \ref{desnow}. This is because the appearance and distribution of snow is similar to that of some types of rain. Based on this observation, we can further infer that our two-stage data distillation method can also be applied to other image reconstruction tasks, such as denoising and image inpainting, under specific parameter settings. These tasks all seek to restore a clean image from the damaged input image, which is damaged in a manner similar to that of some rainy images.

On the other hand, we extend our mission to semantic segmentation to verify the potential value of our network in practical applications. Since rain streaks can blur and block objects, the performance of semantic segmentation will degrade in rainy weather. Figure \ref{segmentation} shows visual results of semantic segmentation by combining with DeepLabv3+ \cite{chen2018encoder}. It is obviously that rain streaks can degrade the performance of DeepLabv3+,  \emph{i.e.}, by missing objects and producing poor segmentation maps. Compared with other methods, our method can remove rain streaks more effectively and deliver better segmentation results along object boundaries.

\section{Conclusion}
We have proposed a two-stage data distillation method for single image rain removal. Instead of using a large amount of paired synthetic data to train a non-robust network, we focus on training a derain network with powerful generalization capabilities using only real rainy images. In the absence of a clean label, we distill knowledge from the input data twice to construct the corresponding soft and hard objectives. Guided by the soft and hard objectives, our derain network can learn to map the input rainy image into a high-quality rain-free image by transferring rain to a high quality clean image to create a more realistic training pair. Experiments verify the superiority of our two-stage data distillation method and also shows the potential of our method to other vision and image restoration tasks.

{\small
	\bibliographystyle{ieee}
	\bibliography{egbib}
}

\end{document}